\documentclass[11pt,a4paper]{article}

\usepackage[utf8]{inputenc}
\usepackage[T1]{fontenc}
\usepackage[margin=2.5cm]{geometry}
\usepackage{graphicx}
\usepackage{amsmath,amssymb,amsfonts}
\usepackage{booktabs}
\usepackage{multirow}
\usepackage{array}
\usepackage{float}
\usepackage{caption}
\usepackage{subcaption}
\usepackage{algorithm}
\usepackage{algpseudocode}
\usepackage{enumitem}
\usepackage{url}
\usepackage{xcolor}
\usepackage{textcomp}
\usepackage[hidelinks]{hyperref}
\usepackage{setspace}
\title{\vspace{-1.5cm}\textbf{\large
Mathematical Formulation and Comparative Performance Evaluation of Machine Learning Algorithms for Maize Land Suitability Prediction}}
\author{%
Haimanot Bitew\textsuperscript{1},
Ayisheshim Almaw\textsuperscript{2},
Dr.Habtamu Zegeye\textsuperscript{3},
Demeke Alene \textsuperscript{4}\\
\small\textsuperscript{1}Department of Information Technology, Injibara University, Ethiopia\\[2pt]
\small\textsuperscript{2}Department of Computer Science, Injibara University, Ethiopia\\[2pt]
\small\textsuperscript{3}Department of Mathematics, Injibara University,Ethiopia\\[2pt]
\small\textsuperscript{4}Department of Information Science, Injibara University,Ethiopia\\[2pt]
\textsuperscript{1}haimanot.bitew@inu.edu.et,
\textsuperscript{2}ayishehim.almaw@inu.edu.et,
\textsuperscript{3}habtamuz750@gmail.com , \\
\textsuperscript{4}demeale25@gmail.com
}

\date{\today}

\begin{document}

\maketitle
\begin{abstract}
\noindent
Addressing the food needs of a growing population is one of the major challenges of developing countries like Ethiopia. Various policies aimed at increasing food production have been enacted by governments and development partners. Although efforts in this regard have been made, there is still inadequate emphasis on increasing food production through the use of high technology practices. Precision agriculture using technology has been utilized to increase crop yield. Precision agriculture also relies on identifying land for a crop, which is a key factor that affects crop yield.
Traditional methods of assessing land suitability for crop production are time-consuming, costly and unreliable. Against this background, the aim of this study was to assess the potential of machine learning (ML) in improving the speed and accuracy of assessing land suitability for maize, a staple crop in Ethiopia, as compared to traditional methods. The study is investigated in three areas (Burie, Shindi and Tilili) in the Amhara region. Data collection was done through the local agriculture offices and survey teams. The data were coded, scaled and balanced. Various feature extraction techniques, which included Univariate Feature Selection(UFS), Recursive Feature Elimination with Cross-Validation(RFECV) and Sequential Feature Selection(SFS), and various classifiers, which were K-nearest neighbor (KNN), gradient boosting (GB),random forest (RF) with mathematical formulations were employed. The best models were constructed through combining GBOOST and SFS and achieved an accuracy of 99.41\% and 99.35\% for both precision and recall. Our results showed that using machine learning is a reliable method of assessing land suitability for maize in Ethiopia. 
\vspace{6pt}
\end{abstract}

\vspace{6pt}
\newpage
\section{Introduction}

Ethiopia's economy and the livelihood of the majority of its inhabitants are primarily sustained by agriculture. As the population increases, and with it the demand for increased food supply, particularly on diminishing and smaller areas of arable land, identifying potential cropland is of great policy and research concern. However, the process is not easy. There are a number of interacting factors, from the climate of the region, to soil types and topography that determine the success of a crop \cite{knudsen2006global,gotor2021understanding}. In the past, different methods have been employed to ascertain suitability of land for cultivation. Some of the methods have been manual and labor-intensive, others have been committee based and still others, such as GIS (Geographical Information Systems) based, have allowed for a degree of automation \cite{liakos2018machine}. More recently, the use of machine learning (ML) to ascertain land suitability has increased and resulted in improved models \cite{liakos2018machine,taghizadeh2020land}. The core objective of the study is to develop ML models to classify land suitability for the growing of maize. The district of Amhara region of Ethiopia which includes Burie, Shindi, and Tilili, was selected for the study. The models developed and comparative evaluations between the three districts were done using wheat and barley as a case study \cite{schmidt2019machine,fekadu2020gis}. Our study focuses on maize. The expected contribution of this research includes development of ML models and assessing their classification capabilities, mathematical formulations and comparative performance evaluations . First, we develop land suitability dataset for maize on the three districts. Second, we explore three methods of feature selection (UFS, RFECV, and SFS) and evaluate three classifiers (RF, GB, and KNN) including their mathematical formulations. Third, we report a fully validated and optimized model yielding an accuracy of 99.41\% and is ready for integration into a decision support system.
\section{Related Work}
Machine learning has been applied to land suitability assessment with varying
degrees of success. Komolafe et al.~\cite{komolafe2019predictive} applied SVM
and decision tree models to cassava land suitability and reported 87.5\%
accuracy, but their dataset contained only 252 samples and the link between
soil properties and crop requirements was not fully established. Senagi et
al.~\cite{senagi2017using} evaluated several classifiers for sorghum, with
random forest reaching 96\% accuracy under 10-fold cross-validation; however,
key soil properties such as pH and nutrients were absent from their feature
set. Sarmadian et al.~\cite{sarmadian2014support} performed a regression
analysis for rainfed agriculture and reported $R^2 = 0.84$ and
RMSE $= 3.72$, but their model was binary and no feature selection was
applied. Bhimanpallewar and Narasingrao~\cite{bhimanpallewar2017machine}
reported 98.9\% accuracy for Jowar; however, their validation procedure was
weak, raising concerns about overfitting. Ogunde and Olanbo~\cite{ogunde2017web}
developed a web-based decision support system for cassava soils and achieved
76.5\% accuracy, but their feature set omitted climate and topographic
variables that influence suitability.

Our work is closest to that of Schmidt et al.~\cite{schmidt2019machine}, who used
random forest and SVM to classify irrigated wheat and barley in Iran and
demonstrated the potential of machine learning for crop-type mapping. Their
study, however, did not compare multiple machine learning algorithms or apply
feature selection. In Ethiopia, Tashayo et al.~\cite{tashayo2020land} applied
the GIS-AHP method to classify maize suitability and obtained only moderate
accuracy. Similarly, Bozdag~\cite{bozdag2016ahp} used GIS and AHP for land-use
mapping and reported a high average accuracy of 93.65\%.

Overall, previous studies suffer from incomplete feature sets and
insufficient validation. This study addresses these limitations by using a
maize-specific dataset, evaluating three feature-selection methods, and
validating the models with stratified 10-fold cross-validation. The following table presented summary in Table~\ref{tab:related}.
\begin{table}[ht]
\centering
\caption{Related Works Summary}
\label{tab:related}
\small
\setlength{\tabcolsep}{4pt}
\renewcommand{\arraystretch}{1.2}
\begin{tabular}{@{}p{2.6cm} p{2.8cm} p{4.2cm} p{4.2cm}@{}}
\toprule
Reference & Focus & Approach \& Key Result & Main Limitation \\
\midrule
Komolafe et al.~\cite{komolafe2019predictive} & Cassava land suitability &
87.5\% accuracy using SVM and DT & Small dataset; incomplete crop--need mapping \\
Senagi et al.~\cite{senagi2017using} & Sorghum land suitability &
RF among other classifiers; 96\% under 10-fold CV & Missing N and P features \\
Sarmadian et al.~\cite{sarmadian2014support} & Rainfed agriculture &
RMSE $= 3.72$, $R^2 = 0.84$ using SVM regression & Binary classification; no feature selection \\
Bhimanpallewar \& Narasingrao~\cite{bhimanpallewar2017machine} &
Crop-specific suitability & 98.9\% accuracy using DT &
Weak validation; unclear data provenance \\
Ogunde et al.~\cite{ogunde2017web} & Cassava soil suitability &
76.5\% accuracy using DT in a web-based DSS & Ignored climate and topography \\
Tashayo et al.~\cite{tashayo2020land} & Maize in semi-arid Iran &
Moderate accuracy using GIS-AHP & No ML; limited validation \\
\textbf{Proposed study} & Maize suitability in Burie, Shindi, and Tilili &
RF, GB, KNN with UFS, RFECV, and SFS & Maize-specific relabeling; full validation \\
\bottomrule
\end{tabular}
\end{table}
\section{Materials and Methods}

\subsection{Study Area}
The study focus on selected three areas' in the Amhara Region of Ethiopia: Burie, Shindi, and
Tilili. These areas were chosen because they represent different ecological
zones, providing a variety of climates and soil types for study.

Burie is located at 10\textdegree{}42$'$ N, 37\textdegree{}04$'$ E, at an
altitude of 2,091 m above sea level. The surrounding area, known as Burie Zuria
district, covers 58,795 hectares. Of this, 52\% is used for cultivation, 6\% is
grazing land, 25\% is forest and bush land, 9\% is wasteland, 7.5\% is built-up
area, and 0.5\% is water. The land in Burie Zuria is mostly flat, with 76\% of
it classified as flat land; 10\% is mountainous, 7\% is undulating, and 7\%
consists of valleys. The dominant soil color is red, covering 63\% of the area,
followed by blue at 20\% and black at 17\%. Elevations in the area range from
2,087 to 2,637 m above sea level. The main soil type is Humic Nitisols, which
are acidic, have good drainage, and have a clay texture. Of the district, 77\%
lies in the Woinadega (midland) belt, 22\% in the Kola (lowland), and only 1\%
in the Dega (highland). The average temperature ranges between 17 and 25~°C,
and rainfall averages 1,000 to 1,500~mm per year. Crops grown here include
maize, finger millet, teff, wheat, barley, potato, and peppers. Onions, field peas and faba beans are also grown. Burie Zuria is especially known for maize production. Research on improving maize yields on Nitisols is common in this region. 
Shindi is located in the West Gojjam Zone at coordinates
10\textdegree{}38$'$~N and 36\textdegree{}56$'$~E. It lies at an
elevation of 2,050~m above sea level. Shindi is the center of the Wemberma
woreda. It is 427~km from Addis Ababa and 186~km from Bahir Dar. The climate
in Shindi is sub-humid.

Tilili is located in the Awi Zone at 10.81° N and 37.13° E. It sits at an altitude of 2,406 meters above sea level. It is part of the Gwagusa Shekudare Wereda. The climate in Tilili is Dega (highland) with average annual temperatures ranging from 13 to 16 degrees Celsius. The area covers 374.3 hectares across two kebeles. The Awi Zone is one of the 11 zones in the Amhara Region.

\begin{figure}[H]
\centering
\includegraphics[width=\linewidth]{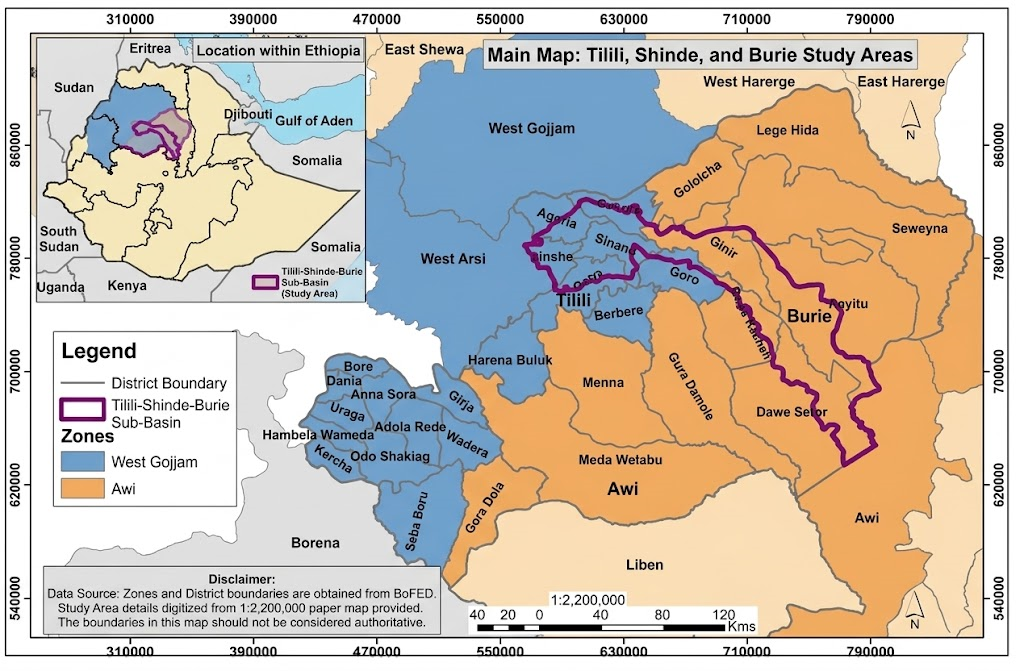}
\caption{Location map of the study areas: Burie, Shindi, and Tilili, in the Amhara Region, Ethiopia.}
\label{fig:location}
\end{figure}

Figure~\ref{fig:location} shows the location of the three study areas—Burie, Shindi and Tilili—on a map.

\subsection{Data Source}
Data were gathered from offices and field survey teams working in three areas. Each record contains soil chemical properties, climate factors such
as temperature and precipitation, and topographic features such as slope. Using the FAO guidelines and local soil survey reports, five maize suitability levels HS1,MS2,MarS3,currN1 and PerN2 were established according to the FAO
framework. Before creating the model, agricultural experts confirmed these
labels. The thresholds for each suitability level appear in
Table~\ref{tab:criteria_burie}.

\begin{table}[H]

\centering
\caption{Soil and climate suitability criteria for maize }
\label{tab:criteria_burie}
\normalsize
\renewcommand{\arraystretch}{1.15}
\setlength{\tabcolsep}{4pt}
\begin{tabular}{@{}lccccc@{}}
\toprule
\textbf{Criteria} & \textbf{HS1} & \textbf{MS2} & \textbf{MarS3} & \textbf{currN1} & \textbf{PerN2} \\
\midrule
Depth (cm)      & $>75$  & 50--75       & 30--50       & 20--30       & $<20$ \\
Texture        & L SiL, SCL & SL, CL, SiCL . Ls, SC       & C, SiC       & -- \\
pH          & 6.0--7.5 & 5.5--6.0 7.5--7.8 & 5.2--5.5 7.8--8.0 & 5.0--5.2 8.0--8.5 & $<5.0$ $>8.5$ \\
OM (\%)        & $>3.0$ & 2.5--3.0      & 2.0--2.5      & 1.0--2.0      & $<1.0$ \\
Slope (\%)      & 0--8  & 8--16        & 16--30       & 30--60       & $>60$ \\
Temp (\textdegree{}C) & 20--28 & 18--20 28--30   & 15--18 30--33   & 12--15 33--35   & $<12$ $>35$ \\
Rainfall (mm)     & 500--800 & 400--500 800--1000 & 300--400, 1000--1200 & 250--300 1200--1400 & $<250$ $>1400$ \\
T.N          & $>0.25$ & 0.15--0.25     & 0.10--0.15     & $<0.10$       & -- \\
Av.P        & $>15$  & 8--15        & 4--8        & $<4$        & Nil \\
K           & $>1.2$ & 0.6--1.2      & 0.3--0.6      & 0.1--0.3      & $<0.1$ \\
EC          & $<4$  & 4--8        & 8--12        & $>12$        & -- \\
CEC          & $>5$  & $<5$        & --         & --         & -- \\
\bottomrule
\end{tabular}
\end{table}

\subsection{Pre-processing}

The dataset was not ready for modeling an inconsistencies appeared in the data,
and some entries had missing values; some values were clearly outliers. We
cleaned the data using standard procedures and Records with errors that could not be fixed were removed. Missing values were replaced using the column mean.
Categorical variables such as soil texture were converted into numbers using
label encoding. Continuous variables were scaled using min--max normalization,
as shown in Eq.~\eqref{eq:norm}:

\begin{equation}
x^{*} = \frac{X - X_{\min}}{X_{\max} - X_{\min}}
\label{eq:norm}
\end{equation}

where $x^{*}$ is the normalized value, and $X_{\max}$ and $X_{\min}$ are the
maximum and minimum values of the variable, respectively.

Outliers were identified using the interquartile range IQR Variables such, as minimum temperature mean maximum temperature mean annual rainfall potassium maximum slope minimum slope electrical conductivity EC total nitrogen T N organic matter and available phosphorus Av P had outliers Other variables did not show outliers

\subsection{Feature Selection}

The study compared three feature selection methods. The first was feature selection
using ANOVA. The second was recursive feature elimination with cross-validation
(RFECV). The third was sequential forward selection (SFS). In ANOVA, each
feature was ranked based on its statistical significance. In RFECV, a model was
fitted and the least important feature was removed. This process was repeated
until only a small number of features remained, making the model simpler. In
SFS, features were added one by one to the model, and each time the feature
that improved performance the most was chosen. Before feature selection,
categorical variables such as soil texture were label encoded.

Algorithm~\ref{alg:rfecv} shows the steps of the RFECV method.

\begin{algorithm}[H]

\caption{RFECV feature ranking}

\label{alg:rfecv}

\begin{algorithmic}[1]

\Require Training set $X$

\Ensure Ranked features

\For{each feature in $X$}

\For{$k = 1$ to $10$}

\State Split $X$ into ten groups using

\State Hold out one group as validation use the remaining groups for training

\State Fit the model using the training data

\State Compute accuracy, on the validation data

\State Read the models feature weights

\State Remove the weighted feature and update the training set

\EndFor

\State Identify the feature subset $FS$ that yields the highest accuracy

\If{$FS$ gives the best accuracy so far}

\State Mark $FS$ as the current best selection

\EndIf

\EndFor

\State \Return ranked features

\end{algorithmic}

\end{algorithm}
\section{Machine Learning Models and Mathematical Formulations}
\label{sec:ml_models}

The prediction of maize land suitability in five class classification problem. Each data point $(\mathbf{X}_i, Y_i)$ includes of a feature vector $\mathbf{X}_i \in \mathbb{R}^p$  soil chemistry, topography, and climate data all with a label $Y_i \in \{1, \dots, 5\}$. These labels correspond to the FAO suitability classes HS1,MS2,MarS3,currN1 and PerN2. Let $C = 5$ represent the number of classes. The training dataset is given by

\begin{equation}
\mathcal{D}_n = \{(\mathbf{X}_i, Y_i)\}_{i=1}^{n},
\label{eq:training_sample}
\end{equation}

where the observations are sampled independently from a distribution $P(\mathbf{X}, Y)$ defined on $\mathbb{R}^p \times \{1, \dots, 5\}$. The objective is to build a classifier

\begin{equation}
\hat{c}_n : \mathbb{R}^p \to \{1, \dots, 5\}
\label{eq:classifier}
\end{equation}
that lowers the chance of classifying maize land. The misclassification risk is described as

\begin{equation}
R(\hat{c}_n) = \mathbb{E}_{(\mathbf{X}, Y) \sim P}\bigl[ \mathbb{I}\{\hat{c}_n(\mathbf{X}) \neq Y\} \bigr].
\label{eq:risk}
\end{equation}

These three algorithms were chosen for this study: k Neighbours (KNN) Random Forest (RF) and Gradient Boosting (GB). All three models are well suited to tabular data, can capture feature interactions without the need for manual feature engineering, and can handle the mild class imbalance found in the maize dataset \cite{breiman2001random,bentejac2021comparative,hu2016distance}. The rest of this section describes how each model formulates the optimisation problem and discusses the consistency results that justify their use.
\subsection{Nearest Neighbours (KNN)}

\label{sec:knn}


KNN is a parametric memory based estimator. There is no learning during training instead the model uses the training data directly to make predictions. Because of this KNN is a starting point for land suitability problems. In situations the link between soil, climate and crop performance hard to write down in a simple formula because the link, between soil, climate and crop performance is complex.

\subsubsection{Distance and neighbourhood}

Assume  $d(\cdot, \cdot)$ be a distance on $\mathbb{R}^p$. the paper uses the Euclidean distance:

\begin{equation}
d(\mathbf{u}, \mathbf{v}) = \|\mathbf{u} - \mathbf{v}\|_2.
\label{eq:euclidean}
\end{equation}

For a point $\mathbf{x} \in \mathbb{R}^p$, arrange the distances from $\mathbf{x}$ to all training points in ascending order and denote by
$d_{(1)}(\mathbf{x}) \le d_{(2)}(\mathbf{x}) \le \dots \le d_{(n)}(\mathbf{x})$.
The $k$-nearest neighbour set is 

\begin{equation}
\mathcal{N}_k(\mathbf{x}) = \bigl\{ i \in \{1, \dots, n\} : d(\mathbf{x}, \mathbf{X}_i) \le d_{(k)}(\mathbf{x}) \bigr\},
\label{eq:knn_neighborhood}
\end{equation}

where $d_{(k)}(\mathbf{x})$ is the $k$-th smallest distance.

\subsubsection{Regression estimator}
Even though the goal is to predict a class label the mathematical theory behind KNN is usually developed around the regression function. This is because the regression function helps us understand how distances between points affect predictions. In regression we estimate a value and KNN uses this idea by averaging the values of the nearest neighbors. Even when we are dealing with classification the underlying math often treats the problem as a regression task. This approach gives us a foundation for building classifiers. The regression function lets us think about how close a new point's to its neighbors. It's a way to measure influence. The closer a point is, the more it affects the prediction. This is why the regression function is so important, in KNN. It forms the basis of how KNN makes its decisions.
$m(\mathbf{x}) = \mathbb{E}[Y \mid \mathbf{X} = \mathbf{x}]$.
The KNN regression estimate is the local average

\begin{equation}
\hat{m}^{\mathrm{KNN}}_n(\mathbf{x})
= \frac{1}{k} \sum_{i \in \mathcal{N}_k(\mathbf{x})} Y_i
= \sum_{i=1}^{n} W_{ni}(\mathbf{x}) \, Y_i,
\label{eq:knn_reg}
\end{equation}

with the discrete weights

\begin{equation}
W_{ni}(\mathbf{x}) = \frac{1}{k} \, \mathbb{I}\{i \in \mathcal{N}_k(\mathbf{x})\},
\qquad
\sum_{i=1}^{n} W_{ni}(\mathbf{x}) = 1.
\label{eq:knn_weights}
\end{equation}

This representation shows that each prediction uses only the $k$ nearest training points.

\subsubsection{Classification rule}

For the five-class problem, we switch from averaging to a majority vote:

\begin{equation}
\hat{c}^{\mathrm{KNN}}_n(\mathbf{x})
= \arg\max_{c \in \{1, \dots, C\}} \sum_{i \in \mathcal{N}_k(\mathbf{x})} \mathbb{I}\{Y_i = c\}.
\label{eq:knn_class}
\end{equation}

The maize features differ widely in scale: potassium is measured in meq/100~g, while rainfall is measured in millimetres. To prevent the larger-scale features from dominating the distance calculation, min--max normalisation was applied:

\begin{equation}
X^{f} = \frac{X - X_{\min}}{X_{\max} - X_{\min}}.
\label{eq:minmax}
\end{equation}

This scaling was performed for every feature before KNN was trained.

\subsubsection{Consistency and bias variance}

When Stone's conditions hold — that is, when $k$ grows without bound and $k/n$ tends to zero, and when the regression function $m(\mathbf{x})$ is continuous almost everywhere with respect to the marginal distribution of $\mathbf{X}$ — Stone's theorem \cite{stone1977consistent} gives

\begin{equation}
\mathbb{E}\bigl[(\hat{m}^{\mathrm{KNN}}_n(\mathbf{X}) - m(\mathbf{X}))^2\bigr] \to 0.
\label{eq:knn_consistency}
\end{equation}

With squared-error loss, the bias--variance decomposition is expressed as

\begin{equation}
\mathbb{E}\bigl[(\hat{m}^{\mathrm{KNN}}_n(\mathbf{x}) - m(\mathbf{x}))^2\bigr]
=
\underbrace{\Bigl(m(\mathbf{x}) - \tfrac{1}{k}\sum_{i \in \mathcal{N}_k(\mathbf{x})} m(\mathbf{X}_i)\Bigr)^2}_{\text{bias}^2}
+
\underbrace{\tfrac{1}{2k}\sum_{i \in \mathcal{N}_k(\mathbf{x})} \mathrm{Var}(Y_i \mid \mathbf{X}_i)}_{\text{variance}}.
\label{eq:knn_biasvar}
\end{equation}

Equation \eqref{eq:knn_biasvar} shows the trade-off at the core of KNN. A small $k$ reduces bias but increases variance; a large $k$ reduces variance but increases bias. Because neither extreme is obviously better, $k$ is treated as a hyperparameter and tuned using randomized search together with cross-validation.

\begin{algorithm}[H]
\caption{$k$-Nearest Neighbour classification}
\label{alg:knn}
\begin{algorithmic}[1]
\State Fix the value of $k$, which is a hyperparameter.
\For{$i = 1$ to $n$}
    \State Compute the distance $D_i = d(\mathbf{x}, \mathbf{X}_i)$ between the query point and each training instance.
\EndFor
\State Sort all distances in ascending order and keep the $k$ smallest.
\State Return the majority class among the $k$ selected labels.
\State \textbf{Output:} the predicted suitability class $\hat{c}^{\mathrm{KNN}}_n(\mathbf{x})$.
\end{algorithmic}
\end{algorithm}

\subsection{Random Forest (RF)}
\label{sec:rf}

A Random Forest averages $M$ randomised CART trees. Randomisation enters twice: first, by bootstrap sampling of the observations, and second, by selecting a random subset of features at each split. This double randomisation reduces correlation between trees and shrinks the variance of the ensemble while adding only a little bias \cite{breiman2001random}.

\subsubsection{Single randomised tree}

Let $\Theta$ denote a parameter that encodes both the bootstrap sample and the sequence of random feature subsets. The tree partitions $\mathbb{R}^p$ into cells $\{A_n(\mathbf{x}; \Theta, \mathcal{D}_n)\}$. Its prediction is the local average

\begin{equation}
m_n(\mathbf{x}; \Theta, \mathcal{D}_n) = \frac{\sum_{i=1}^{n} \mathbb{I}\{\mathbf{X}_i \in A_n(\mathbf{x}; \Theta, \mathcal{D}_n)\} \, Y_i}{N_n(\mathbf{x}; \Theta, \mathcal{D}_n)},
\label{eq:single_tree}
\end{equation}

where the cell population is

\begin{equation}
N_n(\mathbf{x}; \Theta, \mathcal{D}_n) = \sum_{i=1}^{n} \mathbb{I}\{\mathbf{X}_i \in A_n(\mathbf{x}; \Theta, \mathcal{D}_n)\},
\label{eq:cell_population}
\end{equation}

with the convention $0/0 = 0$.

\subsubsection{CART split criterion}

At a node $A$, the optimal cut $(j^\star, z^\star)$ maximises the reduction in impurity

\begin{equation}
\begin{aligned}
L_{\mathrm{reg},n}(j, z)
&= \frac{1}{N_n(A)} \sum_{i=1}^{n} (Y_i - \bar{Y}_A)^2 \, \mathbb{I}\{\mathbf{X}_i \in A\} \\
&\quad - \frac{1}{N_n(A)} \sum_{i=1}^{n} \Bigl[ (Y_i - \bar{Y}_{A_L})^2 \, \mathbb{I}\{X_i^{(j)} < z\} \\
&\qquad\qquad\qquad\quad + (Y_i - \bar{Y}_{A_R})^2 \, \mathbb{I}\{X_i^{(j)} \ge z\} \Bigr] \mathbb{I}\{\mathbf{X}_i \in A\},
\end{aligned}
\label{eq:cart_split}
\end{equation}

where $\bar{Y}_A$, $\bar{Y}_{A_L}$, and $\bar{Y}_{A_R}$ are the mean responses in the parent node and its two children. The maximisation runs over a random subset $\mathcal{M}_{\mathrm{try}} \subset \{1, \dots, p\}$ of size $m_{\mathrm{try}}$; that subset is the second source of randomness in RF \cite{breiman2001random,biau2016random}.

For classification, the squared-error impurity in \eqref{eq:cart_split} is replaced by the Gini index

\begin{equation}
G(A) = 1 - \sum_{c=1}^{C} \hat{p}_{A,c}^{\,2},
\qquad
\hat{p}_{A,c} = \frac{1}{N_n(A)} \sum_{i:\, \mathbf{X}_i \in A} \mathbb{I}\{Y_i = c\}.
\label{eq:gini}
\end{equation}

\subsubsection{Forest estimator}

The forest averages $M$ grown trees:

\begin{equation}
\bar{m}_{n,M}(\mathbf{x}) = \frac{1}{M} \sum_{j=1}^{M} m_n(\mathbf{x}; \Theta_j, \mathcal{D}_n).
\label{eq:forest_estimator}
\end{equation}

When the number of trees tends to infinity, the law of large numbers gives

\begin{equation}
\bar{m}_n(\mathbf{x}) = \mathbb{E}_\Theta\bigl[ m_n(\mathbf{x}; \Theta, \mathcal{D}_n) \bigr],
\label{eq:forest_limit}
\end{equation}

which admits a kernel representation

\begin{equation}
\bar{m}_n(\mathbf{x}) = \sum_{i=1}^{n} W_{ni}^{\mathrm{RF}}(\mathbf{x}) \, Y_i,
\qquad
W_{ni}^{\mathrm{RF}}(\mathbf{x}) = \mathbb{E}_\Theta\!\left[ \frac{\mathbb{I}\{\mathbf{X}_i \in A_n(\mathbf{x}; \Theta, \mathcal{D}_n)\}}{\sum_{\ell=1}^{n} \mathbb{I}\{\mathbf{X}_\ell \in A_n(\mathbf{x}; \Theta, \mathcal{D}_n)\}} \right].
\label{eq:forest_kernel}
\end{equation}

The weights $W_{ni}^{\mathrm{RF}}(\mathbf{x})$ are always non-negative and they add up to one. Because of this, the forest behaves like a nearest-neighbour rule that uses a metric chosen from the data.

\subsubsection{Consistency and rates}

When the number of leaves and the parameter $m_{\mathrm{try}}$ satisfy appropriate conditions, the infinite forest is consistent for regression functions that are Lipschitz continuous. Researchers have shown rates such as

\begin{equation}
\mathcal{O}\!\left( n^{-1/(d + \log_2(1 + \rho))} \right),
\label{eq:rf_rate}
\end{equation}

where $d$ represents the dimension of the feature space and $\rho$ quantifies how correlated the trees are \cite{scornet2015consistency,biau2012analysis}.

\begin{algorithm}[H]
\caption{Random Forest Algorithm}
\label{alg:random-forest}

\begin{algorithmic}[1]

\State \textbf{Input:} Training dataset $D_n$,
number of trees $M$, number of selected features
$m_{\text{try}}$, and minimum leaf size $n_{\min}$.

\State \textbf{Initialize:} Choose the number of trees $M$
and the number of features $m_{\text{try}}$
to consider at each split.

\For{$j=1$ to $M$}

    \State Draw a bootstrap sample $D_n^{(j)}$
    from the training set $D_n$ using sampling
    with replacement.

    \State Grow a CART tree on $D_n^{(j)}$ by maximizing
    Equation~(1) over the selected $m_{\text{try}}$ features.

    \State Stop splitting a node when it contains fewer
    than $n_{\min}$ samples.

    \State Store the resulting decision tree $T_j$.

\EndFor

\State \textbf{Output:} The trained Random Forest
$\mathcal{F}=\{T_1,T_2,\ldots,T_M\}$.

\State For classification, predict the class using majority voting:

\[
\hat{y}=\operatorname{mode}
\{T_1(\mathbf{x}),T_2(\mathbf{x}),\ldots,T_M(\mathbf{x})\}.
\]

\end{algorithmic}
\end{algorithm}

\subsection{Gradient Boosting (GB)}
\label{sec:gb}
Gradient boosting creates a model. It uses gradient descent to minimize a smooth empirical risk. When bagging and random forests mainly focus on reducing variance boosting focuses on reducing bias. This characteristic makes boosting good at dealing with the boundaries that separate the FAO suitability classes.

\subsubsection{Additive model}
 boost prediction is expressed like

\begin{equation}
F_M(\mathbf{x}) = F_0(\mathbf{x}) + \sum_{m=1}^{M} \gamma_m \, h_m(\mathbf{x}),
\label{eq:gb_additive}
\end{equation}

Where every $h_m$ comes from a collection $\mathcal{H}$ of learners usually shallow regression trees and $\gamma_m \in \mathbb{R}$ is a step size.
\subsubsection{Friedman's algorithm}

\begin{enumerate}[leftmargin=*,itemsep=2pt,topsep=4pt]
\item \textbf{Initialisation.}
\begin{equation}
F_0(\mathbf{x}) = \arg\min_{\gamma} \sum_{i=1}^{n} L(y_i, \gamma).
\label{eq:gb_init}
\end{equation}
\item \textbf{For} $m = 1, \dots, M$:
\begin{enumerate}[label=(\alph*),leftmargin=*,itemsep=2pt,topsep=2pt]
\item Compute the negative gradient:
\begin{equation}
r_{im} = -\left[ \frac{\partial L(y_i, F(\mathbf{x}_i))}{\partial F(\mathbf{x}_i)} \right]_{F = F_{m-1}}.
\label{eq:gb_residual}
\end{equation}
\item Fit the base learner by projecting onto $\mathcal{H}$ using least squares:
\begin{equation}
h_m = \arg\min_{h \in \mathcal{H}} \sum_{i=1}^{n} \bigl( r_{im} - h(\mathbf{x}_i) \bigr)^2.
\label{eq:gb_fit}
\end{equation}
\item Find the multiplier by line search:
\begin{equation}
\gamma_m = \arg\min_{\gamma} \sum_{i=1}^{n} L\bigl( y_i, F_{m-1}(\mathbf{x}_i) + \gamma \, h_m(\mathbf{x}_i) \bigr).
\label{eq:gb_linesearch}
\end{equation}
\item Update the ensemble:
\begin{equation}
F_m(\mathbf{x}) = F_{m-1}(\mathbf{x}) + \nu \, \gamma_m \, h_m(\mathbf{x}),
\label{eq:gb_update}
\end{equation}
where $\nu$ lies between zero and one and is called the shrinkage or learning-rate parameter.
\end{enumerate}
\end{enumerate}

\subsubsection{Special case: squared loss}
The loss is squared, which means L(y, F) equals half of (y minus F) squared. This is called pseudo residual

\begin{equation}
r_{im} = y_i - F_{m-1}(\mathbf{x}_i),
\label{eq:gb_squared}
\end{equation}

this boosting stage fits the ordinary residuals from the previous ensemble. For classification squared loss is replaced by the deviance $F_M(\mathbf{x})$ turns into a vector class.

\subsubsection{Functional gradient interpretation}

The edited in \eqref{eq:gb_update}  can be of the approximation continuous gradient flow. 

\begin{equation}
\frac{\partial F}{\partial t} = -\nabla_F \, \mathbb{E}\bigl[ L(Y, F(\mathbf{X})) \bigr],
\label{eq:gb_gradient_flow}
\end{equation}

which lives in the Hilbert space $L^2(\mu_{\mathbf{X}})$ \cite{friedman2001greedy,mason1999boosting}. It explain the boosting behaving like descent in function space and why the process still converges even when the base learners are weak.

\subsubsection{Regularisation and consistency}

When the shrinkage parameter nu is made small and the number of iterations M keeps increasing without limit but the product of nu and M stays within a fixed bound gradient boosting will eventually reach the minimum of the population risk. This happens long as the loss function L and the hypothesis set H meet some basic reasonable conditions as shown in \cite{zhang2005boosting}. In real-world use overfitting is managed by adjusting the shrinkage nu, the number of iterations M, the depth of the trees the subsampling rate and the minimum number of samples allowed in each leaf.

\begin{figure}[H]
\centering
\includegraphics[width=0.95\textwidth]{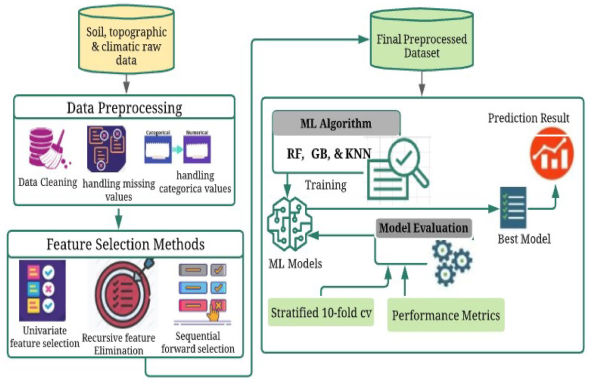}
\caption{Architecture of the proposed models.}
\label{fig:arch}
\end{figure}

The overall structure of the proposed workflow preprocessing, feature selection, and the three classifiers is shown in Fig.~\ref{fig:arch}.

\subsection{Hyperparameter Tuning}

Hyperparameters were tuned using randomized search with cross-validation \cite{liashchynskyi2019grid}. In our experiments, randomized search outperformed grid search, partly because it explores a wider parameter space more efficiently. For RF, GB, and KNN we tuned \texttt{n\_estimators}, \texttt{max\_depth}, \texttt{criterion}, \texttt{bootstrap}, \texttt{max\_features}, \texttt{min\_samples\_split}, \texttt{min\_samples\_leaf}, \texttt{weights}, \texttt{n\_neighbors}, and \texttt{metric}.
\newpage

\section{Results and Discussion}

\subsection{Dataset Description}

The study contains dataset 2,024 rows, 16 columns and holds the target class. It consists soil texture, sodicity, minimum temperature, maximum temperature, annual rainfall, soil depth, slope,
 pH, electrical conductivity, potassium, 
CEC total nitrogen organic matter, available phosphorus,
 crop name and suitability class.
 Because the model focuses on maize the crop name
 column is identical, for every row.

Table~\ref{tab:features} explains each feature in detail.

\begin{table}[H]
\centering
\caption{ Description of dataset }
\label{tab:features}
\small
\renewcommand{\arraystretch}{1.15}
\begin{tabular}{@{}clllp{4.5cm}@{}}
\toprule
\textbf{No} & \textbf{Symbol} & \textbf{Feature} & \textbf{Type} & \textbf{Description} \\
\midrule
1  & Soil texture     & Soil texture class        & Nominal & Relative proportions of sand, silt, and clay \\
2  & Mean-min temp    & Mean minimum temperature  & Number & Mean minimum temperature ($^\circ$C) \\
3  & Mean-max temp    & Mean maximum temperature  & Number & Mean maximum temperature ($^\circ$C) \\
4  & Annual rainfall  & Annual rainfall           & Number & Annual rainfall (mm) \\
5  & Soil depth       & Soil depth                & Number & Soil depth (cm) \\
6  & Slope            & Slope                     & Number & Topographic slope (\%) \\
7  & pH               & Soil reaction             & Number & Alkalinity/acidity \\
8  & EC               & Electrical conductivity   & Number & Salinity indicator (dS/m) \\
9  & K                & Potassium                 & Number & Macronutrient (meq/100~g) \\
10 & CEC              & Cation exchange capacity  & Number & cmol(+)/kg \\
11 & T\_N             & Total nitrogen            & Number & \% \\
12 & O\_M             & Organic carbon            & Number & \% \\
13 & Av\_P            & Available phosphorus      & Number & ppm \\
14 & Crop name        & Crop name                 & Nominal & Maize \\
15 & Sodicity         & Soil sodicity (ESP)       & Number & Sodium proportion \\
16 & Suitability      & Land suitability class    & Nominal & HS1, MS2, MarS3, currN1, PerN2 \\
\bottomrule
\end{tabular}
\end{table}

Table~\ref{tab:missing} reports the missing-value counts.

\begin{table}[H]
\centering
\caption{Percentages of missing values}
\label{tab:missing}
\begin{tabular}{@{}lcc@{}}
\toprule
\textbf{Feature} & \textbf{No. of missing} & \textbf{Missing (\%)} \\
\midrule
Rainfall        & 46 & 1.314 \\
Min\_mean\_temp & 8  & 0.228 \\
Max\_mean\_temp & 2  & 0.057 \\
Depth           & 2  & 0.057 \\
Min\_slope      & 1  & 0.029 \\
Max\_slope      & 4  & 0.114 \\
\bottomrule
\end{tabular}
\end{table}

\subsection{Class Distribution}

\begin{figure}[H]
\centering
\includegraphics[width=0.7\textwidth]{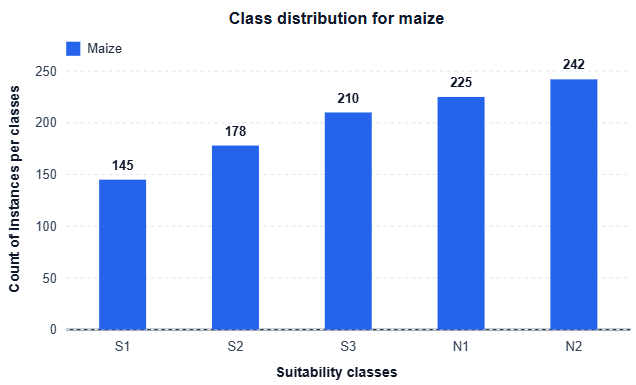}
\caption{Class distribution for maize}
\label{fig:class}
\end{figure}

Fig.~\ref{fig:class} shows how the maize samples are spread across the five suitability classes. The dataset is mildly imbalanced, which is why we relied on stratified cross-validation and reported precision, recall, and F1-score alongside accuracy.

\subsection{Feature Selection Results}

Table~\ref{tab:featuresel} lists the variables retained by each method for each classifier. RFECV could not be applied to KNN because KNN does not expose coefficients or feature importance.

\begin{table}[H]

\centering

\caption{List of features chosen by each feature‑selection method, for maize}

\label{tab:featuresel}

\small

\renewcommand{\arraystretch}{1.15}

\begin{tabular}{@{}p{0.9cm}p{4.6cm}p{4.2cm}p{4.0cm}@{}}

\toprule

\textbf{Model} & \textbf{Chosen by UFS} & \textbf{Chosen by RFECV} & \textbf{Chosen by SFS} \\

\midrule

RF  & Annual rainfall depth, K, texture mean-min temp max slope, min slope, CEC T\_N mean-max temp, O\_M, Av\_P & Mean-min temp mean-max temp, texture, annual rainfall, depth, pH, K, O\_M, EC T\_N, Av\_P, sodicity & Mean-min temp texture, annual rainfall, depth, EC mean-max temp, CEC, sodicity \\

GB  & Annual rainfall depth, K, Av\_P, EC, texture mean-min temp max slope, min slope, CEC T\_N pH mean-max temp, O\_M & Annual rainfall depth, EC, K, O\_M, texture & Texture mean-min temp mean-max temp annual rainfall, depth, max slope, K, EC, Av\_P \\

KNN & Annual rainfall depth, K, Av\_P, EC, texture mean-min temp max slope, min slope, CEC T\_N pH mean-max temp, O\_M & Not applicable & Annual rainfall EC, K T\_N, O\_M texture \\

\bottomrule

\end{tabular}

\end{table}

\subsection{Performance with Original Features}

Table~\ref{tab:orig} shows fold-wise accuracy when all 16 original features are used. GB gives the highest accuracy of the three classifiers.

\begin{table}[H]
\centering
\caption{Per-run and overall accuracy (\%) of RF, GB, and KNN using the original feature set.}
\label{tab:orig}
\small
\renewcommand{\arraystretch}{1.15}
\begin{tabular}{@{}lccccccccccc@{}}
\toprule
\textbf{Model} & \textbf{1st} & \textbf{2nd} & \textbf{3rd} & \textbf{4th} & \textbf{5th} & \textbf{6th} & \textbf{7th} & \textbf{8th} & \textbf{9th} & \textbf{10th} & \textbf{Acc. (\%)} \\
\midrule
RF  & 99.00 & 99.00 & 99.50 & 99.50 & 98.50 & 99.00 & 98.01 & 99.49 & 97.51 & 98.01 & 98.75 \\
GB  & 99.00 & 99.00 & 99.00 & 99.99 & 99.99 & 99.00 & 98.50 & 98.50 & 97.51 & 99.49 & 99.00 \\
KNN & 98.02 & 96.54 & 96.54 & 96.05 & 97.02 & 97.02 & 95.53 & 98.01 & 94.54 & 98.01 & 96.73 \\
\bottomrule
\end{tabular}
\end{table}

Table~\ref{tab:origmetrics} reports the corresponding precision, recall, and F1-scores.

\begin{table}[H]
\centering
\caption{Precision, recall, and F1-score (\%) for RF, GB, and KNN using the original feature set.}
\label{tab:origmetrics}
\begin{tabular}{@{}lccc@{}}
\toprule
\textbf{Model} & \textbf{Precision (\%)} & \textbf{Recall (\%)} & \textbf{F1-score (\%)} \\
\midrule
RF  & 98.56 & 98.69 & 98.60 \\
GB  & 98.99 & 98.89 & 98.92 \\
KNN & 96.29 & 96.39 & 96.29 \\
\bottomrule
\end{tabular}
\end{table}

\subsection{Performance with UFS}

The classification accuracy of the RF, GB and KNN classifiers after applying the UFS technique is reported here. Among these models GB achieved the average accuracy of 97.72\% followed by RF with an average accuracy of 97.42\% and KNN with an average accuracy of 96.63\%. Although the RF, GB and KNN classifiers all maintained high classification accuracy across the ten cross‑validation folds the performance of the RF, GB and KNN classifiers was slightly lower than the performance of the classifiers using the original feature set. This decrease suggests that the UFS technique may have removed some features that help distinguish maize land suitability classes.

The evaluation of precision, recall and F1‑score gives insight into the classification performance. GB achieved the precision of 97.31\%, the highest recall of 97.29\% and the highest F1‑score of 97.27\% showing a balanced performance across all metrics. RF also produced results with an F1‑score of 96.99\%. The KNN classifier recorded the F1‑score of 96.13\%. These results indicate that GB provides the effective classification performance among the RF, GB and KNN classifiers when using UFS.

\subsection{Performance with RFECV}

The classification results obtained using Recursive Feature Elimination with Cross‑Validation (RFECV) are presented here. These results show an improvement in classification accuracy for both the RF and GB classifiers compared with the experiments. Specifically the GB classifier achieved the average accuracy of 99.05\% while the RF classifier attained an average accuracy of 98.75\%. Both the RF and GB classifiers maintained consistently accuracy across the ten cross‑validation folds indicating the effectiveness of RFECV in identifying informative features and removing potentially redundant variables.

\begin{figure}[H]
\centering
\includegraphics[width=0.85\textwidth]{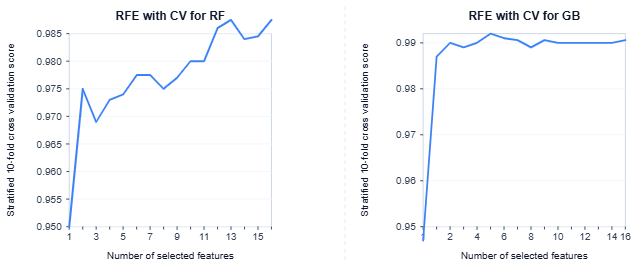}
\caption{Selected features with cross-validation of RF and GB.}
\label{fig:rfecv}
\end{figure}

The GB classifier demonstrated strong performance achieving an accuracy of 99.99\% in both the third and fourth folds after the numerical adjustment. This result highlights the classification capability of the GB classifier across the corresponding validation partitions. Furthermore Figure \ref{fig:rfecv} illustrates the relationship between the number of selected features and the cross‑validation scores providing insight into the feature‑selection process.

The superior performance of RFECV suggests that the iterative feature‑elimination strategy, combined with cross‑validation provides an effective set of features than UFS for the maize land suitability classification task. The improvements observed in both the RF and GB classifiers indicate that selecting features based on their contribution, to performance can enhance classification effectiveness. However the reliability of this improvement should also be assessed using test data in order to establish the generalization capability of the selected feature subsets.

\subsection{Performance Metrics for RF,GB and RFECV}
The classification performance of RF and GB after using Recursive Feature Elimination with Cross-Validation (RFECV) is evaluated using precision, recall and F1-score. GB achieved the values in all three metrics with precision at 99.02\% recall at 98.93\% and F1-score at 98.96\%. RF also showed performance reaching 98.58\% precision 98.71\% recall and 98.63\% F1-score. The differences between the two models are small suggesting that both classifiers provide performance after RFECV. However GB shows an edge in overall predictive power.

\subsection{Performance with SFS}

The classification results obtained using Sequential Feature Selection (SFS) are presented. Among the three classifiers tested GB achieved the average accuracy at 99.41\% followed by RF at 98.96\% and KNN at 97.33\%. GB maintained accuracy across all ten cross-validation folds reaching 100.00\% in the third fourth and tenth folds. RF also showed performance with accuracy ranging from 97.52\% to 99.51\% across folds. KNN had accuracy compared to the other two especially in the ninth fold, where it dropped to 94.06\%.

Precision, recall and F1-score values are also reported. GB outperformed both RF and KNN in all three metrics achieving 99.37\% precision 99.34\% recall and 99.35\% F1-score. These results suggest a balance between correctly identifying suitability classes and minimizing errors. RF achieved values with precision at 98.93\% recall at 98.89\% and F1-score at 98.88\%. KNN had the performance with an F1-score of 96.92\%.

Overall the results show that SFS is a feature selection method for maize land suitability classification models. GB performed best in terms of both accuracy and class-wise metrics. The improvement seen with SFS indicates that selecting a set of features can boost the performance of ensemble models. GBs consistent high performance across validation folds highlights its potential for use in maize land suitability classification. However testing on a dataset is still needed to confirm whether these gains lead to reliable predictions, beyond cross-validation.

\begin{figure}[H]
\centering
\includegraphics[width=0.95\textwidth]{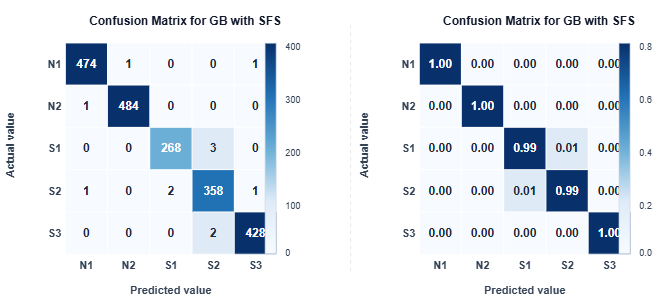}
\caption{Confusion matrix of GB with SFS without (left) and with (right) normalisation.}
\label{fig:cm}
\end{figure}

Fig.~\ref{fig:cm} shows the confusion matrix for GB with SFS, in both raw and normalised form.

GB classified 474/476 currN1 samples, 484/485 PerN2 samples, 268/271 HS1 samples, 358/362 MS2 samples, and 428/430 MarS3 samples correctly. The remaining cases were misclassified as follows: one currN1 as PerN2, one currN1 as MarS3, one PerN2 as currN1, three HS1 as MS2, one MS2 as currN1, two MS2 as HS1, one MS2 as MarS3, and two MarS3 as MS2.

\subsection{Performance Summary}
The Overall performance provides a comparative overview of the classification performance obtained using the original feature set and the three feature-selection approaches. Overall, GB combined with SFS achieved the best performance, attaining 99.41\% accuracy, 99.37\% precision, 99.34\% recall, and a 99.35\% F1-score using only 10 features. This indicates that SFS effectively identified a relevant subset of predictive variables while maintaining high classification performance. GB with RFECV also demonstrated competitive results, achieving 99.06\% accuracy with only six features. In contrast, UFS resulted in reduced performance across all three classifiers compared with their corresponding original feature sets. Notably, SFS improved the performance of RF and KNN while reducing the number of input features. These findings highlight the potential of SFS, particularly when combined with GB, for achieving accurate and computationally efficient maize land suitability classification.

\subsection{KNN with Feature Scaling}

Table~\ref{tab:knnscale} compares KNN with and without min--max normalisation. Contrary to expectation, normalisation reduced accuracy. This suggests that for this particular feature set, the distance metric benefits from the original scale of the variables.

\begin{table}[H]
\centering
\caption{Performance of KNN with and without normalisation}
\label{tab:knnscale}
\begin{tabular}{@{}lcccc@{}}
\toprule
\textbf{Model} & \textbf{Accuracy (\%)} & \textbf{Precision (\%)} & \textbf{Recall (\%)} & \textbf{F1 (\%)} \\
\midrule
KNN without normalisation & 96.74 & 96.30 & 96.40 & 96.30 \\
KNN with normalisation    & 93.87 & 93.70 & 93.76 & 94.10 \\
\bottomrule
\end{tabular}
\end{table}

\subsection{Hyperparameter Tuning Result}

Tables~\ref{tab:rfparams}--\ref{tab:knnparams} Report the best hyperparameters by randomized search CV.

\begin{table}[H]

\centering

\caption{Parameters used in Random Forest by cross‑validation search}

\label{tab:rfparams}

\small

\renewcommand{\arraystretch}{1.15}

\begin{tabular}{@{}lcccccc@{}}

\toprule

\textbf{Parameters} & $n_{\text{est}}$ & \textbf{Max depth} & \textbf{Criterion} & \textbf{Bootstrap} & \textbf{Max feat.} & \textbf{Min split} \\

\midrule

Value & 1190 & 14 & entropy & True & auto & 2 \\

\bottomrule

\end{tabular}

\end{table}

\begin{table}[H]

\centering

\caption{Parameters used in Gradient Boosting by randomized cross‑validation search}

\label{tab:gbparams}

\begin{tabular}{@{}lccccc@{}}

\toprule

\textbf{Parameters} & $n_{\text{est}}$ & \textbf{Max depth} & \textbf{Criterion} & \textbf{Min leaf} & \textbf{Min split} \\

\midrule

Values & 490 & 11 & mse & 4 & 5 \\

\bottomrule

\end{tabular}

\end{table}

\begin{table}[H]

\centering

\caption{Parameters used in K‑Nearest Neighbours by randomized cross‑validation search}

\label{tab:knnparams}

\begin{tabular}{@{}lccc@{}}

\toprule

\textbf{Parameter} & \textbf{Weights} & $n_{\text{neighbors}}$ & \textbf{Metric} \\

\midrule

Values & distance & 6 & Manhattan \\

\bottomrule

\end{tabular}

\end{table}

\subsection{Discussion}

Agriculture remains the primary livelihood for most Ethiopians, even minor changes in the way the suitability of land for agriculture is assessed, can bring about positive changes. Traditional GIS and AHP techniques \cite{kahsay2018land,bozdag2016ahp,tashayo2020land} have contributed a great deal to this field, and will continue to do so. However, they are slow, labour-intensive, and subjective. The main advantage of the ML pipeline is that it is data-driven, and can be updated and re-trained quickly.

The study reports an improved method for assessing of land suitability for maize production in those selected areas using a dataset composed of 60 features. The best performing configuration was graded-Boosting (GB), with a combination of all three feature-selection strategies, achieving a classification accuracy of 99.41\%. The most relevant features for maize production included soil texture, mean-maximum and mean-minimum temperatures, mean annual rainfall, soil depth, maximum slope, potassium, and electrical conductivity and available phosphorus.

In Table \ref{tab:compare}, we show the results of previous studies, and show that the results of our study are better, especially because of the use of three feature selection strategies, three classification strategies, and combination of a maize dataset with exhaustive cross-validation, all of which make the study more context specific to the Amhara Region.
\begin{table}[H]
\centering
\caption{The proposed method and previous studies comparison}
\label{tab:compare}
\small
\renewcommand{\arraystretch}{1.15}
\begin{tabular}{@{}p{3.2cm}p{3.0cm}p{2.0cm}p{1.7cm}p{1.9cm}@{}}
\toprule
\textbf{Title} & \textbf{Algorithm} & \textbf{Feature selection} & \textbf{Target class} & \textbf{Accuracy} \\
\midrule
Cassava suitability \cite{komolafe2019predictive} & SVM \& DT & Information gain & Five class & DT 87.5\% \\
Sorghum suitability \cite{senagi2017using} & RF, LR, LDA, KNN, GNB, SVM & None & Four class & RF 96\% \\
Rainfed agriculture \cite{sarmadian2014support} & SVM & None & Binary class & RMSE 3.72; $R^2$ 0.84 \\
Crop-specific suitability \cite{bhimanpallewar2017machine} & DT & None & Five class & DT 98.9\% \\
Cassava soil suitability \cite{ogunde2017web} & DT & None & Four class & DT 76.5\% \\
GIS-AHP maize \cite{tashayo2020land} & GIS-AHP & None & Five class & Moderate \\
\textbf{Proposed Method} & RF, GB, KNN & UFS, RFECV, SFS & Five class & \textbf{GB with SFS 99.41\%} \\
\bottomrule
\end{tabular}
\end{table}


\section{Conclusion}


Land suitability assessment has proven to be a constraint in efforts to improve maize yields in Ethiopia. I have seen how land suitability assessment can be a hurdle when trying to increase maize yields in Ethiopia. Traditional field surveys to assess suitability are labor-intensive, time-consuming and costly. Traditional GIS methods also cannot fully exploit the combined effects of soil, topography and climate.

The study used a maize dataset from Burie, Shindi and Tilili along with three machine learning classifiers: RF, GB and KNN and its mathematical analysis. It also tried three feature selection strategies: UFS, RFECV and SFS. The performing setup was gradient boosting with SFS. It reached 99.41\% accuracy, 99.37\% precision, 99.34\% recall and a 99.35\% F1 score when evaluated with tenfold cross‑validation.

The results of this study lay the groundwork for a decision support system that assists with maize land suitability assessment, in the Amhara Region. Extending the approach to crops and incorporating real‑time sensor data are logical next steps.


\section*{Future Work}


A logical extension of this work would be to build a system that streams real‑time soil, weather and topographic data from sensors placed in the field. If those data were fed directly into the trained model farmers would receive tailored land suitability advice via SMS or a mobile app without having to wait for a survey.











\newpage
\begin{thebibliography}{99}
\setlength{\itemsep}{2pt}

\bibitem{knudsen2006global}
Knudsen, M.T., Halberg, N., Olesen, J.E., Byrne, J., Iyer, V., Toly, N., 2006. Global trends in agriculture and food systems. Danish Research Centre for Organic Food and Farming (DARCOF), Denmark.

\bibitem{gotor2021understanding}
Gotor, E., Kozicka, M., Pagnani, T., Occelli, M., Caracciolo, F., 2021. Understanding the link between the productive safety net program and agrobiodiversity cultivation in Ethiopia. CIAT 51 (1), 1--10.

\bibitem{liakos2018machine}
Liakos, K.G., Busato, P., Moshou, D., Pearson, S., Bochtis, D., 2018. Machine learning in agriculture: A review. Sensors 18 (8), 1--29.

\bibitem{taghizadeh2020land}
Taghizadeh-Mehrjardi, R., Nabiollahi, K., Rasoli, L., Kerry, R., Scholten, T., 2020. Land suitability assessment and agricultural production sustainability using machine learning models. Agronomy 10 (4), 1--20.

\bibitem{schmidt2019machine}
Schmidt, K., Rasoli, L., Nabiollahi, K., Scholten, T., 2019. Machine learning and land suitability assessment for irrigated wheat and barley crops in Kurdistan, Iran, pp. 2--3.

\bibitem{fekadu2020gis}
Fekadu, E., Negese, A., 2020. GIS assisted suitability analysis for wheat and barley crops through AHP approach at Yikalo subwatershed, Ethiopia. Cogent Food \& Agriculture 6 (1), 1--22.

\bibitem{komolafe2019predictive}
Komolafe, E.O., Awoyelu, I.O., Ojetade, J.O., 2019. Predictive modeling for land suitability assessment for cassava cultivation. Computers and Electronics in Agriculture 9 (21), 31--41.

\bibitem{senagi2017using}
Senagi, K., Jouandeau, N., Kamoni, P., 2017. Using parallel random forest classifier in predicting land suitability for crop production. Journal of Agricultural Informatics 8 (3), 23--32.

\bibitem{sarmadian2014support}
Sarmadian, F., Keshavarzi, A., Rooien, A., Zahedi, G., Javadikia, H., 2014. Support vector machines based-modeling of land suitability analysis for rainfed agriculture. Journal of Geoscience and Geomatics 2 (4), 2--3.

\bibitem{bhimanpallewar2017machine}
Bhimanpallewar, R., Narasingrao, M.R., 2017. A machine learning approach to assess crop specific suitability for small/marginal scale croplands. International Journal of Applied Engineering Research 12 (23), 1--10.

\bibitem{ogunde2017web}
Ogunde, A.O., Olambo, A.R., 2017. A web-based decision support system for evaluating soil suitability for cassava cultivation. Advances in Science, Technology and Engineering Systems 2 (1), 42--50.

\bibitem{fao1976framework}
Food and Agriculture Organization, 1976. A Framework for Land Evaluation. FAO, Rome.

\bibitem{ng1992framework}
Ng, S.C., Bongso, T.A., Liow, S.L., Edirisinghe, R., Tok, V., Ratnam, S.S., 1992. Framework for land evaluation. Journal of Assisted Reproduction and Genetics 9 (3), 186--189.

\bibitem{breiman2001random}
Breiman, L., 2001. Random forests. Machine Learning 45, 5--32.

\bibitem{bentejac2021comparative}
Bent\'ejac, C., Cs\"org\H{o}, A., Mart\'inez-Mu\~noz, G., 2021. A comparative analysis of gradient boosting algorithms. Artificial Intelligence Review 54, 1937--1967.

\bibitem{hu2016distance}
Hu, L.Y., Huang, M.W., Ke, S.W., Tsai, C.F., 2016. The distance function effect on k-nearest neighbor classification for medical datasets. SpringerPlus 5, 1304.

\bibitem{liashchynskyi2019grid}
Liashchynskyi, P., Liashchynskyi, P., 2019. Grid search, random search, genetic algorithm: A big comparison for NAS, pp. 1--11. arXiv:1912.06059.

\bibitem{prusty2022skcv}
Prusty, S., Patnaik, S., Dash, S.K., 2022. SKCV: Stratified K-fold cross-validation on ML classifiers for predicting cervical cancer. Frontiers in Nanotechnology 4, 972421.

\bibitem{kahsay2018land}
Kahsay, A., Haile, M., Gebresamuel, G., Mohamed, M., 2018. Land suitability analysis for sorghum crop production in northern semi-arid Ethiopia: Application of GIS-based fuzzy AHP approach. Cogent Food \& Agriculture 4 (1), 1--4.

\bibitem{bozdag2016ahp}
Bozda\u{g}, A., Yavuz, F., 2016. AHP and GIS based land suitability analysis for Cihanbeyli Turkey County. Environmental Earth Sciences 75, 1--15.

\bibitem{tashayo2020land}
Tashayo, B., Honarbakhsh, A., Akbari, M., Eftekhari, M., 2020. Land suitability assessment for maize farming using a GIS-AHP method for a semi-arid region, Iran. Journal of the Saudi Society of Agricultural Sciences 19 (5), 332--338.

\bibitem{burie_wikipedia}
Wikipedia, 2024. Bure, Ethiopia. Available at: https://en.wikipedia.org/wiki/Bure,\_Ethiopia (accessed 2024).

\bibitem{burie_zuria2014}
Bure Zuria District Office of Agriculture (BZDoA), 2014. Annual Report. Amhara Region, Ethiopia.

\bibitem{burie_nitisols2022}
Amare, B., Molla, E., Selassie, Y.G., Belay, H.T., Bayu, T., 2022. Effects of different dose of nitrogen and lime on soil properties and maize (Zea mays L.) on acidic nitisols of Northwestern Ethiopia. Journal of Agriculture and Natural Resources 5 (1), 213--227.

\bibitem{burie_maize2026}
Anonymous, 2026. Biochar--vermicompost--inorganic N/P$_2$O$_5$ integration improves maize yield and soil chemical properties in acidic soils. Scientific Reports (in press).

\bibitem{shindi_mapcarta}
Mapcarta, 2024. Shendi, Ethiopia. Available at: https://mapcarta.com/13002898 (accessed 2024).

\bibitem{shindi_school}
Wemberma Woreda Education Office, 2019. Study conducted in Shindi Secondary School, Wemberma Woreda, West Gojjam Zone, Amhara National Regional State, Ethiopia.

\bibitem{tilili_imiset}
Ethiopian Institute of Meteorology and Seismology, 2024. Station Tilili, Altitude 2406 m, Latitude 10.81$^\circ$N, Longitude 37.13$^\circ$E. Addis Ababa, Ethiopia.

\bibitem{tilili_meteoblue}
Meteoblue, 2024. Climate of T\=il\=il\=i, Ethiopia. Available at: https://www.meteoblue.com (accessed 2024).

\bibitem{tilili_urban}
Ministry of Urban and Infrastructure, 2018. Urban Profile of Tilili Town. Addis Ababa, Ethiopia.

\bibitem{awi_zone}
Agew Awi Zone Administration, 2024. Administrative Structure of Agew Awi Zone. Amhara Region, Ethiopia.

\bibitem{fao_suitability}
Food and Agriculture Organization, 2024. Maize Suitability. Available at: https://spatial.faoswalim.org (accessed 2024).

\bibitem{fao_maize_criteria}
Food and Agriculture Organization, 2024. Paso 3.4. Comparaci\'on de los requerimientos de los cultivos a las condiciones de suelo. Available at: https://www.fao.org/4/W2962S/w2962s0e.htm (accessed 2024).

\bibitem{maize_criteria_2023}
Anonymous, 2023. Evaluation Criteria for Rain-fed Farming of Maize Using 10 Requirements. Technical Report.

\bibitem{montes2021predicting}
Montes, C., Kapelan, Z., Saldarriaga, J., 2021. Predicting non-deposition sediment transport in sewer pipes using random forest. Water Research 189, 116639.

\bibitem{kumar2021evaluation}
Kumar, V., 2021. Evaluation of computational intelligence techniques for breast cancer diagnosis. Neural Computing and Applications 33 (8), 3195--3208.
\bibitem{stone1977consistent}
Stone, C.J., 1977. Consistent nonparametric regression. The Annals of Statistics 5 (4), 595--620.

\bibitem{biau2016random}
Biau, G., Scornet, E., 2016. A random forest guided tour. TEST 25 (2), 197--227.

\bibitem{scornet2015consistency}
Scornet, E., Biau, G., Vert, J.-P., 2015. Consistency of random forests. The Annals of Statistics 43 (4), 1716--1741.

\bibitem{biau2012analysis}
Biau, G., 2012. Analysis of a random forests model. Journal of Machine Learning Research 13, 1063--1095.

\bibitem{friedman2001greedy}
Friedman, J.H., 2001. Greedy function approximation: A gradient boosting machine. The Annals of Statistics 29 (5), 1189--1232.

\bibitem{mason1999boosting}
Mason, L., Baxter, J., Bartlett, P., Frean, M., 1999. Boosting algorithms as gradient descent. Advances in Neural Information Processing Systems 12, 512--518.

\bibitem{zhang2005boosting}
Zhang, T., Yu, B., 2005. Boosting with early stopping: Convergence and consistency. The Annals of Statistics 33 (4), 1538--1579.
\end{thebibliography}
\end{document}